\documentclass[conference]{IEEEtran}
\IEEEoverridecommandlockouts
\usepackage{cite}
\usepackage{amsmath,amssymb,amsfonts}
\usepackage{algorithmic}
\usepackage{graphicx}
\usepackage{textcomp}
\usepackage{xcolor}
\def\BibTeX{{\rm B\kern-.05em{\sc i\kern-.025em b}\kern-.08em
    T\kern-.1667em\lower.7ex\hbox{E}\kern-.125emX}}
\begin{document}

\title{Towards Greener Nights: Exploring AI-Driven Solutions for Light Pollution Management\\
}

\author{\IEEEauthorblockN{Paras Varshney}
\IEEEauthorblockA{\textit{Mechanical and Industrial Engineering} \\
\textit{Northeastern University}\\
varshney.p@northeastern.edu}
\and
\IEEEauthorblockN{Niral Desai}
\IEEEauthorblockA{\textit{Mechanical and Industrial Engineering} \\
\textit{Northeastern University}\\
desai.niral1@northeastern.edu}
\and
\IEEEauthorblockN{Uzair Ahmed}
\IEEEauthorblockA{Professor, Khoury College of CS \\
\textit{Northeastern University}\\
u.ahmad@northeastern.edu}

}

\maketitle

\begin{abstract}
This research endeavors to address the pervasive issue of light pollution through an interdisciplinary approach, leveraging data science and machine learning techniques. By analyzing extensive datasets and research findings, we aim to develop predictive models capable of estimating the degree of sky glow observed in various locations and times. Our research seeks to inform evidence-based interventions and promote responsible outdoor lighting practices to mitigate the adverse impacts of light pollution on ecosystems, energy consumption, and human well-being.

\end{abstract}


\section{Introduction}
The burgeoning presence of artificial light at night has ushered in an era where the celestial wonders that once adorned the night sky are increasingly obscured from view, leaving millions bereft of the opportunity to marvel at the grandeur of the Milky Way. According to the International Dark-Sky Association, this pervasive phenomenon, also known as light pollution, extends far beyond mere inconvenience and manifests as a complex issue with serious implications for human health, environmental sustainability, energy conservation, and aesthetic appreciation.

Light pollution encompasses an array of adverse effects, including sky glow, glare, light trespass, and clutter, each contributing to the degradation of our nocturnal environment and impeding our ability to fully appreciate the night sky. Sky glow, characterized by the brightening of the night sky over inhabited areas, not only hampers astronomical observations but also disrupts the natural order of the nocturnal ecosystem. Glare, the sensation produced by excessive luminance within the visual field, poses risks to human safety and visual comfort, diminishing visibility and increasing the likelihood of accidents. Light trespass, wherein light falls where it is neither intended nor wanted, encroaches upon private spaces and disrupts the natural rhythms of wildlife. Additionally, clutter, the proliferation of bright, confusing, and excessive groupings of light sources, contributes to urban sky glow and exacerbates visual pollution in urban environments.

The root of the problem lies in the widespread and indiscriminate use of outdoor lighting that is inefficient, poorly targeted, and inadequately shielded, resulting in the wasteful emission of light into the sky. As cities and urban areas continue to expand, the nighttime landscape is increasingly dominated by artificial illumination, with the natural darkness of the night sky becoming a rare commodity. The pervasive orange hue that characterizes many urban nightscapes stands as a stark testament to the ubiquity of light pollution, obscuring the celestial panorama and eroding our connection to the cosmos.

Addressing the challenges posed by light pollution necessitates a paradigm shift in outdoor lighting practices, guided by principles of efficiency, environmental stewardship, and responsible illumination. By adopting strategies to optimize visibility, minimize energy consumption, reduce glare, and mitigate light trespass, communities can not only preserve the integrity of the night sky but also enhance public safety, conserve natural resources, and promote the well-being of both humans and wildlife. Through concerted efforts to raise awareness, implement regulations, and embrace innovative lighting technologies, we can aspire to reclaim the majesty of the night sky and ensure that future generations have the opportunity to gaze upon the stars with wonder and awe.

In 1879, Thomas Edison’s pioneering incandescent light bulbs cast their glow upon a New York street, marking the genesis of the modern era of electric lighting as mentioned by Chepesiuk R. \cite{b1} Since that transformative moment, the world has been inundated with the radiance of artificial illumination. From powerful street lamps illuminating thoroughfares to the dazzling luminescence of sports arenas visible for miles around, the pervasive presence of electric light has reshaped the nocturnal landscape. According to the International Dark-Sky Association (IDA) based in Tucson, Arizona, the sprawling metropolis of Los Angeles casts a luminous sky glow visible from airborne vantage points over 200 miles away. In many urban centers worldwide, the spectacle of stargazing has become confined to the confines of planetariums, with the natural brilliance of the Milky Way obscured by the luminous veil of urban sky glow.

\section{Related Studies}
\subsection{Scientific Evidences from Nature}

Artificial illumination has ushered in a plethora of societal benefits, extending the boundaries of the productive day and offering enhanced opportunities for both work and leisure activities reliant on illumination. However, the ubiquity of electric lighting does not inherently denote detriment. Yet, amidst this radiance, lies a burgeoning issue – light pollution. When outdoor lighting becomes inefficient, intrusive, and superfluous, it manifests as light pollution, a concern increasingly recognized by environmentalists, naturalists, and medical researchers as one of the most rapidly proliferating forms of environmental degradation. Mounting scientific evidence suggests that light pollution exerts profound and lasting effects on both human and wildlife health.

Light pollution:
\begin{itemize}

\item is a risk factor for cancer to people
\item is a disturbance to animals
\item is a waste of energy, resources, and money
\item is a security risk to your property and obscures our view of the wondrous night sky.
\end{itemize}

\subsection{Effects due to Ubanization}

The thresholds at which nuisance light transmutes into a health hazard remain a topic of fervent investigation. Richard Stevens, a distinguished cancer epidemiologist at the University of Connecticut Health Center, emphasizes the pivotal role of light exposure on circadian rhythm disruption, particularly in environments inundated with artificial illumination such as Manhattan, Las Vegas or Los Angeles as shown in Fig.~\ref{fig_los_angeles_night_sky} below. As humanity grapples with the implications of excessive nocturnal light exposure, studies like "The First World Atlas of the Artificial Night Sky Brightness" published in the Monthly Notices of the Royal Astronomical Society have shed light on the alarming extent of light pollution. The report reveals that a significant portion of the global populace, including two-thirds of the U.S. population and over half of the European population, has forfeited the ability to behold the Milky Way with the unaided eye. Moreover, a substantial portion of the world's inhabitants resides in regions where artificial sky brightness exceeds the threshold for light-polluted status set by the International Astronomical Union.

\begin{figure}[htbp]
\centering
\includegraphics[width=0.4\textwidth]{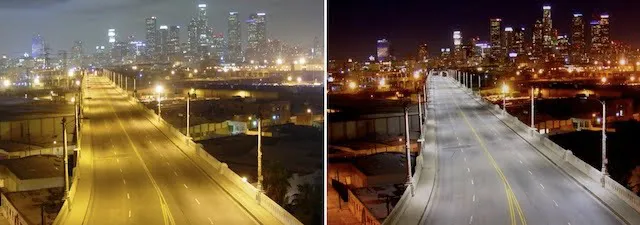}
\caption{The 6th Street Bridge in Los Angeles before (left) and after (right) the LED conversion (Bureau of Street Lighting)}
\label{fig_los_angeles_night_sky}
\end{figure}

\subsection{Effects on our Circadian Cycle}

Growing evidence suggests that exposure to artificial nighttime light can disrupt circadian and neuroendocrine physiology in humans \cite{b1}, potentially accelerating tumor growth and contributing to a spectrum of health disorders, including breast cancer. Paolo Sassone-Corsi, a leading expert in circadian biology, underscores the pivotal role of the circadian clock in regulating physiological processes and its link to various medical conditions. As research in this field intensifies, the urgent need to address the adverse health impacts of light pollution becomes increasingly apparent.

\subsection{LED a major cause for Light Pollution}

Artificial lighting at night has become a significant driver of the escalating light pollution crisis, as elucidated by Kyba et al. \cite{b2}. Through an extensive analysis of data collected by citizen scientists spanning over a decade, the researchers have uncovered a concerning trend: a 10\% annual rise in the sky background's luminance due to artificial light, obscuring even the faintest stars in the night sky. This surge, largely fueled by the widespread adoption of LED technology, poses a substantial challenge for satellite-based monitoring, as current satellite detectors are unable to capture the blue light emitted by LEDs. The findings underscore the pressing need for advanced satellite technologies capable of detecting light pollution across different spectral bands, as highlighted by Fabio Falchi et al. \cite{b3}. Moreover, they emphasize the imperative to combat light pollution comprehensively, given its multifaceted adverse impacts on ecosystems, cultural heritage, and human health.

\begin{figure}[htbp]
\centering
\includegraphics[width=0.4\textwidth]{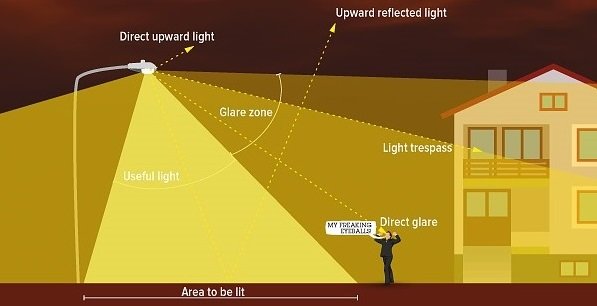}
\caption{The image illustrating different components of light pollution. Courtesy Anezka Gocova, in The Night Issue (2013) \cite{b4}}
\label{fig_components_of_night_light}
\end{figure}

Indeed, a significant portion of outdoor lighting employed during the night suffers from inefficiency, excessive brightness, inadequate targeting, insufficient shielding, and often, needless deployment as shown in Fig.~\ref{fig_components_of_night_light}. This illumination, along with the electricity expended to generate it, is squandered as it disperses into the sky rather than effectively illuminating desired objects and spaces as discussed by Yadav et al. in \cite{b5} 

\begin{itemize}
\item \textbf{Glare} -- Glare refers to excessive brightness that causes visual discomfort or even temporary vision impairment. It occurs when lighting fixtures emit light that is too intense or poorly directed, leading to harsh contrasts and difficulty seeing clearly.

\item \textbf{Skyglow} -- Skyglow is the brightening of the night sky over inhabited areas caused by artificial lighting. It results from light pollution, where excessive and poorly controlled outdoor lighting scatters into the atmosphere, obscuring the view of stars and other celestial objects.

\item \textbf{Light trespass} -- Light trespass occurs when artificial light spills beyond its intended area, encroaching onto neighboring properties, public spaces, or into windows of homes and buildings where it is not needed or desired. This can disrupt sleep patterns, affect wildlife behavior, and diminish the enjoyment of natural darkness.

\item \textbf{Clutter} -- Clutter refers to the presence of numerous bright, confusing, and excessive groupings of light sources, often in urban environments. It can create visual confusion, making it difficult to distinguish important details or navigate effectively. Cluttered lighting environments can be overwhelming and contribute to a sense of unease or discomfort.
\end{itemize}

\begin{figure}[htbp]
\centering
\includegraphics[width=0.4\textwidth]{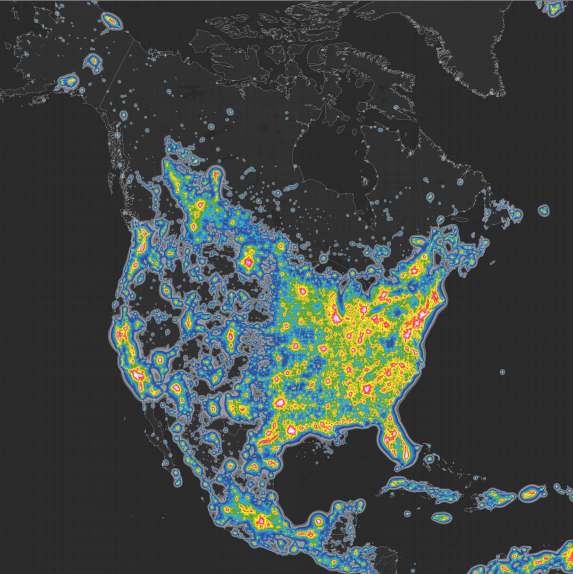}
\caption{Map of North America’s artificial sky brightness, as a ratio to the natural sky brightness. (Falchi et al, 2016) \cite{b3}}
\label{fig_north_america_night_light}
\end{figure}

In a recent investigation published in Science Advances, a collaborative effort by global researchers has yielded the most intricate cartographic representation of light pollution to date. Their findings paint a stark picture: the majestic sight of the Milky Way is now beyond the visual reach of approximately one-third of the global population, with staggering proportions of 60 percent of Europeans and 80 percent of Americans deprived of this celestial spectacle, as shown in Fig.~\ref{fig_north_america_night_light}. The pervasive glow of artificial city lights has engendered a ubiquitous phenomenon known as "skyglow," permanently shrouding the nocturnal heavens and eclipsing our once-unobstructed view of the stars. The accompanying map delineates the extent of artificial sky brightness across North America, providing a nuanced insight into the escalating encroachment of light pollution on the natural nocturnal environment.

\section{Preparing the Dataset}

In our research endeavor, we meticulously curated a dataset by amalgamating infographics and demographic data sourced from the esteemed Globe at Night foundation. Since 2006, this foundation has been steadfast in documenting nocturnal sky conditions, providing a rich repository of observations. Leveraging advancements in technology, particularly the SQM-LE meter by Unihedron, we extended this initiative through the GaN-MN project, establishing a global network for monitoring night sky brightness. This comprehensive dataset encompasses observations spanning over a decade, offering insights into the spatial and temporal dynamics of light pollution.

By integrating GaN datasets spanning from 2006 to 2020, we synthesized a cohesive dataset capturing a myriad of variables elucidating the nuances of nocturnal luminance. This amalgamation ensures a comprehensive representation of global trends in light pollution over an extensive timeframe, facilitating robust analyses and informed decision-making. Noteworthy attributes include geographical coordinates of observers, temporal markers delineating date and time of observation, elevation data, sensor specifications, among others, as delineated below. These variables collectively contribute to a rich tapestry of information, enabling nuanced exploration of factors influencing nocturnal luminance patterns. The focal variable of interest, limiting magnitude, serves as a pivotal metric quantifying the degree of luminosity present in the night sky, constituting the target variable for predictive modeling endeavors. By elucidating variations in limiting magnitude across different spatial and temporal contexts, we aim to glean insights into the drivers of light pollution and formulate strategies for its mitigation and management.

\subsection{Base Training dataset}
\begin{itemize}
    \item \textbf{id}: The unique ID for each data entry.
    \item \textbf{time}: The time at which the data was collected.
    \item \textbf{time\_zone}: The regional timezone of the observation site.
    \item \textbf{country}: Country in which the data was collected.
    \item \textbf{latitude}: The latitude for the location of observation.
    \item \textbf{longitude}: The longitude for the location of observation.
    \item \textbf{elevation\_m}: Elevation from mean sea level (in meters) for the observation site.
    \item \textbf{type}: The type of sensor used for the measurements.
    \item \textbf{sensor\_reading}: The reading of the sensor.
    \item \textbf{clouds}: Category of cloud cover in the sky.
    \item \textbf{constellation}: Constellation formed in the sky as seen by the reporting object at observation location.
    \item \textbf{comment\_1}: Comment for the sky visibility.
    \item \textbf{comment\_2}: Comment for the observation location.
    \item \textbf{limiting\_magnitude}: The scale on which the stars are visible in the night sky. [Target Feature]
\end{itemize}

Upon conducting qualitative analysis, a notable observation emerged indicating a strong correlation between the intensity of nocturnal light and the population density of the corresponding region. This pivotal insight prompted us to augment our dataset with census data delineating regional population demographics. By incorporating this supplementary dataset, we aimed to elucidate the intricate interplay between human habitation patterns and nocturnal luminance levels. The inclusion of population data serves as a crucial adjunct to our research efforts, facilitating a more nuanced understanding of the socio-demographic factors underpinning variations in light pollution.

\subsection{Population dataset (additional)}

\begin{itemize}
    \item \textbf{Country Name}: The country in which the data was collected.
    \item \textbf{year}: Population for the given country for that year 2006 to 2020.
\end{itemize}

\section{Approach}
The first step in our exploratory data analysis (EDA) journey was carefully putting together a large dataset that included infographics and demographic data from the reputable Globe at Night foundation. 

\begin{figure}[htbp]
\centering
\includegraphics[width=0.45\textwidth]{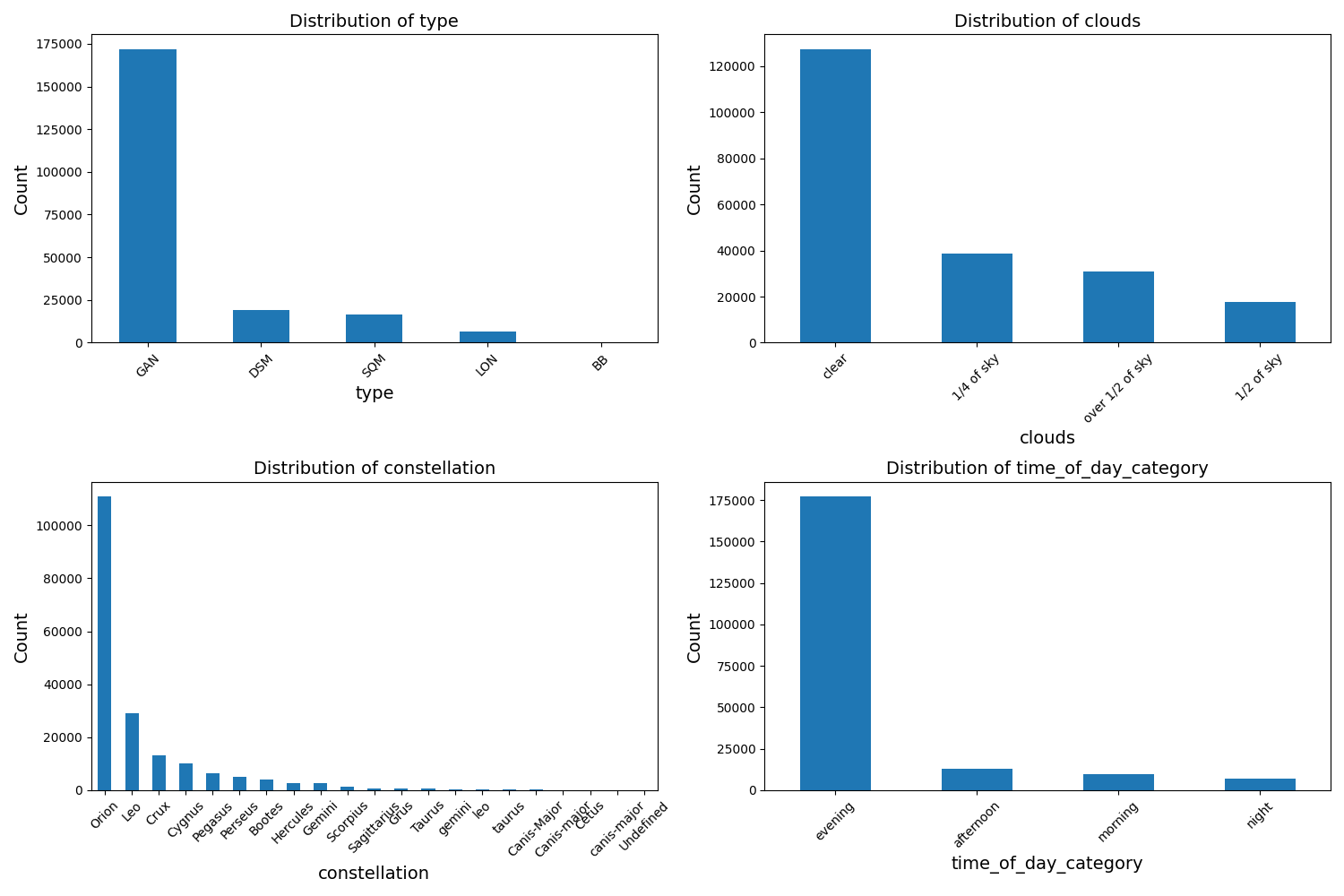}
\caption{Average annual change of various features in contrast of limiting magnitude}
\label{fig_value_counts}
\end{figure}

\subsection{Understanding the Data}

An analysis of the distribution of categorical variables yields significant insights about the dataset. The 'type' column is primarily composed of GAN, accounting for 80.1\% of the entries. Following GAN, the other types listed are DSM, SQM, LON, and BB. The majority of the sky is characterized by clear circumstances, accounting for 59.4\%, while the remaining portion is occupied by different weather conditions. The column labeled 'constellation' is mostly dominated by Orion, accounting for 41.0\% of the entries. It is then followed by Leo, Crux, and various other constellations. The evening category is the most prevalent in terms of 'time\_of\_day\_category,' accounting for 82.7\% of the total. It is followed by the afternoon, morning, and night categories. Gaining insight into these distributions illuminates common patterns within the data, providing valuable information for later studies and interpretations.

\subsection{Statistical Analysis}

Upon examining the dataset, which consists of 214,626 rows, distinct patterns in missing values are evident across different variables. The sensor\_reading column is significantly lacking in data, with 177,691 missing entries, accounting for approximately 82.8\% of the total observations. This signifies a notable disparity in the documented sensor measurements, which could potentially affect the thoroughness of the dataset's analysis. Additionally, the columns comment\_1 and comment\_2 have a significant number of missing values, representing 42.9\% and 48.0\% of the dataset, respectively. The absence of data in the comments sections could hinder the thorough interpretation of observations or events related to the recorded measurements. In addition, the feature time\_since\_last\_purchase only includes missing values, indicating either a lack of relevant data or a specific situation where such information is not useful. In addition, the elevation\_m attribute has a very small number of missing values, accounting for only 0.0014\% of the dataset. On the other hand, variables like constellation and limiting\_magnitude have a substantial amount of missing data, with proportions of 12.1\% and 8.0\% respectively. Additional examination of the characteristics and consequences of missing data in particular attributes is necessary to guarantee the soundness and dependability of subsequent analyses and findings.

\begin{figure}[htbp]
\centering
\includegraphics[width=0.45\textwidth]{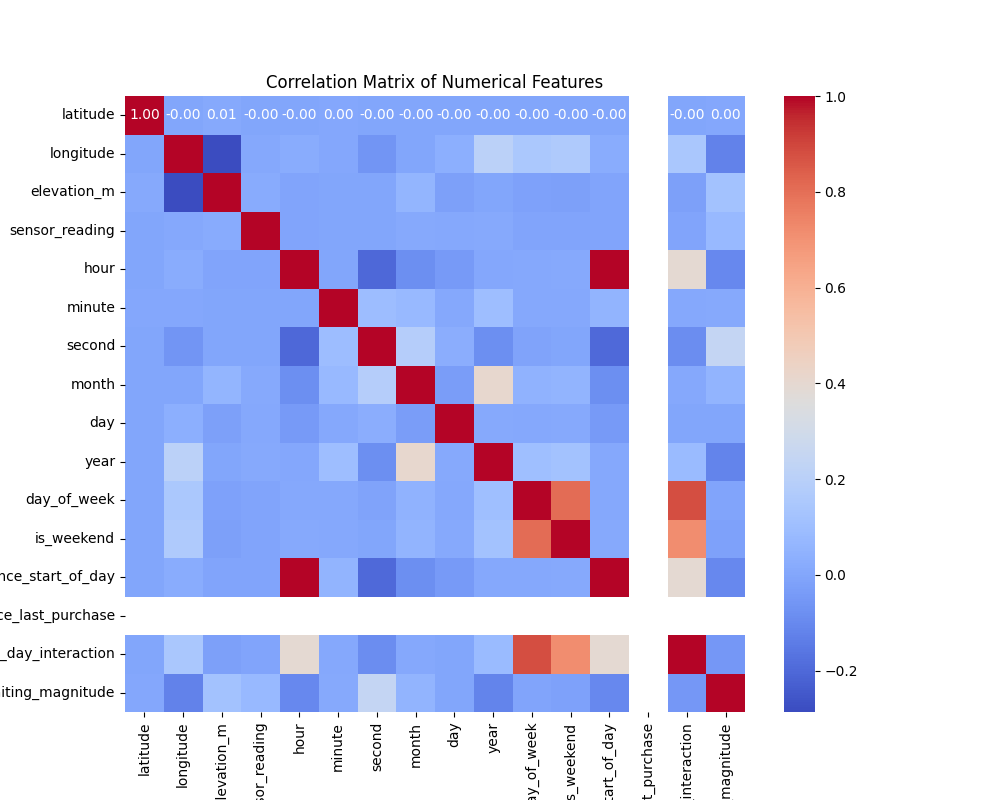}
\caption{Numerical Features Correlations with Target}
\label{fig_correlation_map}
\end{figure}

\subsection{Correlations: Feature Mappings}

The correlation matrix analysis provides valuable insights into the correlations between different numerical features and the target variable, limiting magnitude as shown in Fig.~\ref{fig_correlation_map}. Significantly, there is a moderate positive correlation (0.12) between the year and limiting magnitude, indicating a progressive rise in limiting magnitude as time progresses. Moreover, there is a slight negative correlation (-0.12) between longitude and limiting magnitude, suggesting that specific geographical areas may provide improved vision of celestial objects. Remarkably, there exists a moderate positive correlation (0.25) between the second component of time and limiting magnitude, suggesting the possibility of temporal patterns that affect the visibility of celestial objects. These findings highlight the intricate relationship between temporal, geographical, and astronomical aspects in determining the limiting magnitude. Further research is needed to understand the underlying mechanisms.

\begin{figure}[htbp]
\centering
\includegraphics[width=0.45\textwidth]{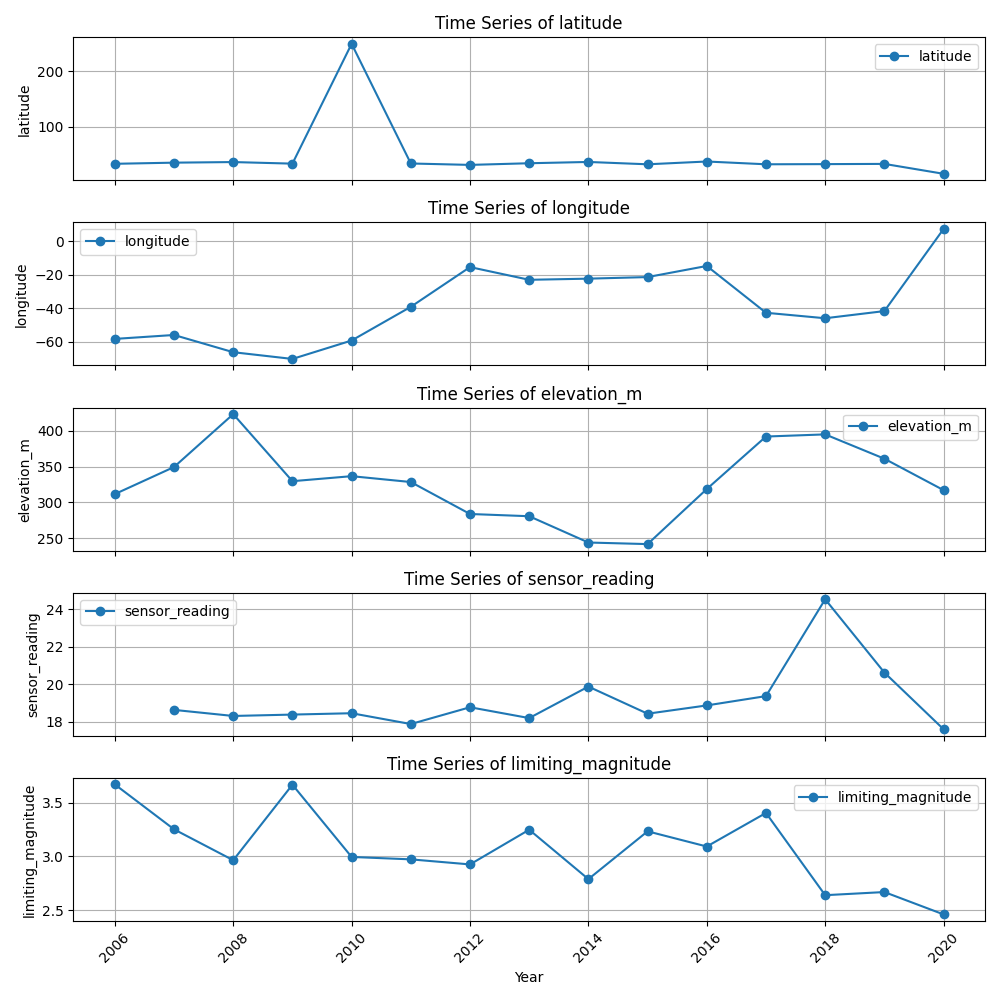}
\caption{Average annual change of various features in contrast of limiting magnitude}
\label{fig_line_charts_comparison}
\end{figure}

\subsection{Trend Analysis}

To find trends and outliers, visual exploration tools like histograms, box plots, and scatter plots were used. One interesting result was that there was a strong link between the amount of light at night and the number of people living in an area. This led us to add census data to our dataset to help us learn more about how socio-demographic factors affect changes in light pollution. This in-depth Exploratory Data Analysis set the stage for later modeling efforts that tried to predict and lessen the effects of light pollution.

In the course of our investigation, we conducted an exhaustive analysis by employing line charts to scrutinize and contrast the annual trends of various numerical attributes. The aim of this investigation was to uncover insights regarding the annual evolution of these qualities and to identify any noteworthy patterns or trends. By aggregating data and calculating annual means, we were able to attain a comprehensive understanding of the patterns and trends demonstrated by these numerical attributes. The line charts served as a visual aid to highlight the fluctuations and recurring trends in each attribute over time, providing a clear and intuitive illustration of the data as shown in Fig.~\ref{fig_line_charts_comparison}. The objective of this analysis was to detect any significant alterations or correlations that might offer valuable insights for determining feature engineering requirements or the selection of modeling strategies in related fields. The findings of our research offer valuable insights into the temporal variations of these quantitative attributes and deepen our understanding of their annual trends and patterns.

\section{Architecture and Data Modelling}

Our modeling approach began with preprocessing the time series data, a crucial step in ensuring its compatibility with subsequent analyses. We standardized the data to address any differences in magnitude across various features, assuring fairness in subsequent procedures. We employed the capabilities of k-nearest neighbors (KNN) to investigate the temporal and spatial aspects of our dataset, specifically examining parameters such as time, timezone, latitude, and longitude. Through the utilization of the K-nearest neighbors (KNN) algorithm as very deeply analyzed in \cite{b10}, our objective was to effectively capture the inherent connections among these variables, so enhancing our dataset with significant contextual insights. This method enabled the development of nearest neighbor mean features, which acted as descriptive indicators of the surrounding environment of each data point, so improving our comprehension of the spatial-temporal patterns within the dataset.

\begin{figure}[htbp]
\centering
\includegraphics[width=0.45\textwidth]{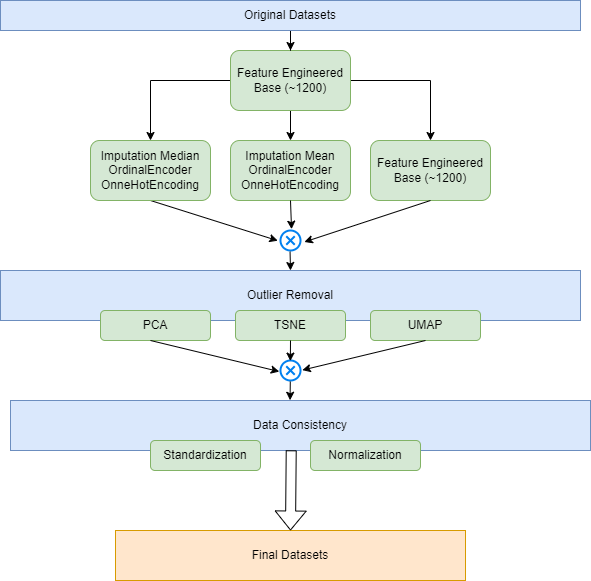}
\caption{Feature Engineering, Outlier Removal and Data Normalization}
\label{fig_fe_and_data_processing}
\end{figure}

Subsequently, we utilized a pre-trained DeBERTa v3 model as described by He, Pengcheng et al. in the paper \cite{b11} that was specifically fine-tuned for our objective. The neural network architecture, widely recognized for its efficacy in natural language processing applications, was modified to include our categorical variables: nation, kind, clouds, constellation, and textual columns (comments\_1 and comments\_2). Using the knowledge gained from a large collection of text data, we obtained complex representations that capture the meaning of each categorical variable. The embeddings successfully captured the intricate linkages and semantic meanings present in the categorical data, allowing our model to understand subtle patterns and important associations necessary for predictive modeling.

\begin{figure}[htbp]
\centering
\includegraphics[width=0.45\textwidth]{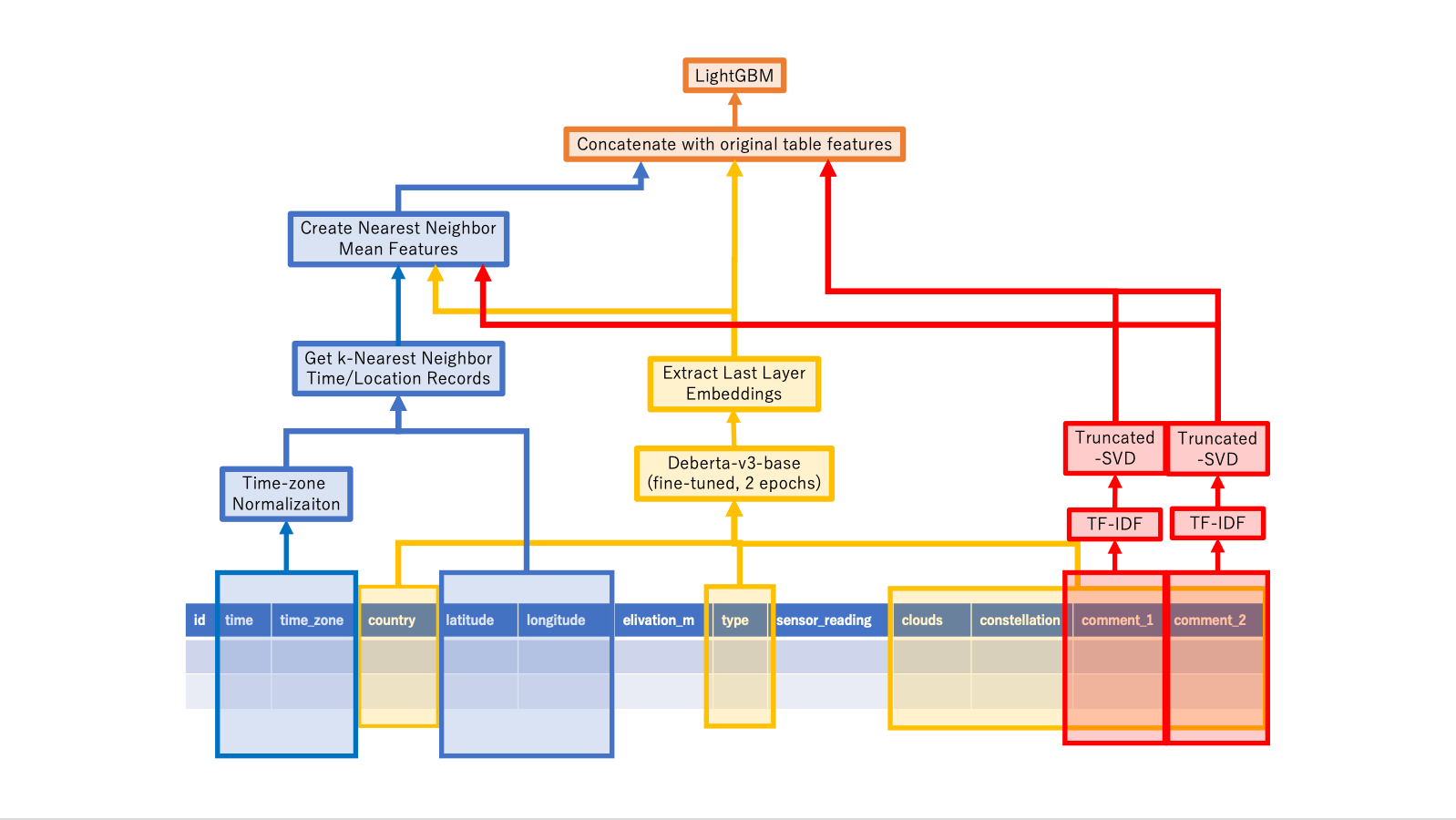}
\caption{Data Modelling Architecture}
\label{fig_architecture_modelling}
\end{figure}

In order to optimize the utilization of the textual data contained in the comments\_1 and comments\_2 columns, we implemented a two-step methodology. Initially, we employed term frequency-inverse document frequency (TF-IDF) as highlighed by Das in his paper \cite{b12} to measure the importance of words in each comment, taking into consideration their relevance in the overall dataset. Afterwards, we utilized truncated singular value decomposition (SVD) to decrease the dimensionality of the TF-IDF matrix while retaining the most informative latent features as shown in Fig.~\ref{fig_fe_and_data_processing}. This approach converted the textual data into a succinct yet useful representation, making it easier to include into our predictive modeling pipeline. Our goal was to improve our model's capacity to understand the subtle meaning in the comments\_1 and comments\_2 columns by reducing the amount of text and representing it in a smaller space. This would make our model better at making predictions.

\subsection{Cross Validation}

Our cross-validation (CV) strategy involved using a random 5-fold CV approach to assess the reliability of our models. Then we also implemented a 10 fold CV to stabilize results and contrast the model stability with the 5 fold CV. This approach was both simple and successful. With a dataset containing 214,626 rows, this strategy effectively evaluated the performance of the model in a fair and accurate manner across various subsets of the data. By including each fold, all observations were guaranteed to be part of both the training and validation sets at least once. This approach reduces bias and provides accurate estimations of the model's performance. Although there were some inconsistencies between the CV scores and leaderboard (LB) scores, which were likely caused by temporal variations in the data distribution, our CV approach yielded valuable insights into the stability and generalization ability of the model. This helped us make informed decisions when selecting the model and tuning the hyperparameters.

\subsection{Ensemble Models}

\begin{figure}[htbp]
\centering
\includegraphics[width=0.45\textwidth]{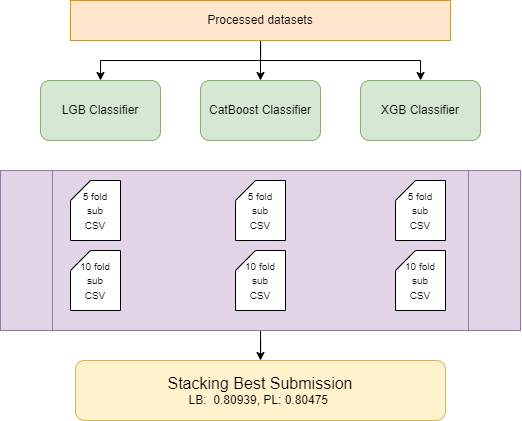}
\caption{Training and Ensembling Architecture}
\label{fig_ensembling_strategy}
\end{figure}

When developing our ensemble strategy, we strategically combined various modeling methodologies to take advantage of their unique strengths and minimize any limitations they may have as shown in Fig.~\ref{fig_ensembling_strategy}. By utilizing the insights gained from our analysis, which considered each row separately and utilized stratified K-fold cross-validation on the target variable, we integrated several preprocessing approaches like outlier clipping and feature engineering to improve the performance of the model. In addition, we incorporated text processing techniques such as TF-IDF with dimensionality reduction using PCA, as well as ensemble methods utilizing XGBoost, LightGBM, and RandomForest as described in \cite{b8} and \cite{b9}. In our ensembling enhanced our strategy, highlighting the importance of a well-balanced weight distribution to maximize predictive performance. By employing a thorough ensemble method, we were able to effectively utilize the combined capabilities of many models and preprocessing techniques. This resulted in the development of a final model that is both resilient and reliable, with improved predictive power and the capacity to generalize well.

\subsection{Evaluation}

Evaluation Metrics: The evaluation metric for this competition is the Mean F1-Score, a widely used measure in information retrieval. The F1 score combines precision and recall, where precision represents the ratio of true positive predictions to the total predicted positives, and recall represents the ratio of true positive predictions to the total actual positives.

\[ \text{F1 Score} = 2 \times \frac{\text{Precision} \times \text{Recall}}{\text{Precision} + \text{Recall}} \]

The Mean F1-Score, also known as the micro-averaged F1-score, is calculated as the harmonic mean of precision and recall, emphasizing balanced performance across both metrics. A higher F1 score indicates better model performance in terms of both precision and recall. In this competition, the final F1 score achieved was 0.81883.

\section{Conclusion}

To summarize, our process of modeling involved a methodical examination of several methodologies and algorithms, guided by a thorough assessment of their performance and appropriateness for our dataset. Through a thorough comparison of CatBoost and LightGBM, we obtained valuable insights into their individual strengths and weaknesses. As a result, we made the decision to use LightGBM as our primary model due to its exceptional computational efficiency and accuracy as shown in table 1. By combining insights from individual model performances and preprocessing techniques, our ensemble strategy allowed us to take advantage of the synergies between multiple approaches, resulting in a strong and scalable predictive model which scored a 0.81883 on the leaderboard and 0.81793 on the public leaderboard. 

\begin{table}[htbp]
\caption{F1-score evaluation metric results}
\begin{center}
\begin{tabular}{|l|c|}
\hline
Model            & F1 Score    \\
\hline
CatBoost         & 0.7925       \\
LightGBM         & 0.8042       \\
DeBerta v3       & 0.7802      \\
Ensemble (Mean)  & 0.8121       \\
Ensemble (Opt)   & 0.8188       \\
\hline
\end{tabular}
\label{tab1}
\end{center}
\end{table}

By employing cross-validation of 5 folds, we have ensured the dependability and capacity for generalization of our models, even in the face of difficulties arising from temporal inconsistencies in the dataset. In summary, our trip highlights the significance of careful testing and assessment in creating accurate prediction models. It emphasizes the use of many methodologies and ensembling strategies to get the best possible performance.

\section*{Acknowledgment}

We extend our heartfelt appreciation to Professor Ahmad Uzair, part-time lecturer at Khoury College of Computer Science and his teaching assistants, for their exceptional guidance and encouragement during the Machine Learning course (CS 6140). His expertise and motivation have been instrumental in our journey, shaping our understanding and passion for the subject. We are deeply grateful for his invaluable support and mentorship.

The code of this project is available at \cite{b13} which is a private repository so please request collaboration rights. If you want to access it please send us an email.






\end{document}